\documentclass[letterpaper, 10 pt, conference]{ieeeconf}  

\IEEEoverridecommandlockouts                              

\usepackage{epsfig} 
\usepackage{mathptmx} 
\usepackage{times} 
\usepackage{amsmath} 
\usepackage{amssymb}  

\usepackage{makeidx}
\usepackage{graphicx}
\usepackage[caption=false, font=footnotesize]{subfig}
\usepackage{amsmath,amssymb} 
\usepackage{color}
\usepackage[rgb]{xcolor}    
\usepackage{tikz}
\usetikzlibrary{calc}
\usepackage{cite}           
\usepackage{verbatim} 		
\usepackage{import}			
\graphicspath{{img/}}	
\usepackage{multirow}       
\usepackage{booktabs}       

\definecolor{daiCoolGrey} {RGB}{230,230,230}
\definecolor{daiCoolGrey1}{RGB}{68,68,68}
\definecolor{daiCoolGrey2}{RGB}{112,112,112}
\definecolor{daiCoolGrey3}{RGB}{158,158,158}
\definecolor{daiCoolGrey4}{RGB}{200,200,200}
\definecolor{daiPetrol} {RGB}{0,103,127}
\definecolor{daiPetrol1}{RGB}{0,51,60}
\definecolor{daiPetrol2}{RGB}{0,67,85}
\definecolor{daiPetrol3}{RGB}{0,86,106}
\definecolor{daiPetrol4}{RGB}{0,122,147}
\definecolor{daiPetrol5}{RGB}{80,151,171}
\definecolor{daiPetrol6}{RGB}{121,174,191}
\definecolor{daiPetrol7}{RGB}{166,202,216}
\definecolor{daiDeepRed} {RGB}{114,23,12}
\definecolor{daiDeepRed1}{RGB}{68,14,7}
\definecolor{daiDeepRed2}{RGB}{90,19,10}
\definecolor{daiDeepRed3}{RGB}{159,25,36}
\definecolor{daiDeepRed4}{RGB}{255,0,0}
\definecolor{daiDeepRed5}{RGB}{140,70,60}
\definecolor{daiDeepRed6}{RGB}{170,115,110}
\definecolor{daiDeepRed7}{RGB}{200,160,160}
\definecolor{daiDeepRed8}{RGB}{230,210,210}
\definecolor{daiOrange} {RGB}{230,145,35}
\definecolor{daiOrange1}{RGB}{235,165,80}
\definecolor{daiOrange2}{RGB}{240,190,125}
\definecolor{daiOrange3}{RGB}{245,210,170}
\definecolor{daiOrange4}{RGB}{250,230,210}

\definecolor{daiGreen} {RGB}{110,160,70}
\definecolor{daiGreen1}{RGB}{140,180,110}
\definecolor{daiGreen2}{RGB}{170,200,145}
\definecolor{daiGreen3}{RGB}{200,200,180}
\definecolor{daiGreen4}{RGB}{225,235,220}

\definecolor{color0}{named}{daiPetrol}
\definecolor{color1}{named}{daiDeepRed}
\definecolor{color2}{named}{daiOrange}
\definecolor{color3}{named}{daiGreen}
\definecolor{color4}{named}{daiCoolGrey2}

\definecolor{boxplot_whisker}{named}{black}
\definecolor{boxplot_median}{named}{daiPetrol}
\definecolor{boxplot_frame}{named}{black}
\definecolor{boxplot_fill}{named}{white}
\definecolor{boxplot_outlier}{named}{daiDeepRed}

\newcommand*\rot{\rotatebox{90}}
\newcommand{\vlpKnnMeanIoURoad}{86.88}

\newcommand{\vlpKnnMeanIoUChamber}{96.40}
\newcommand{\vlpSvmMeanIoUChamber}{97.14}
\newcommand{\scalaKnnMeanIoUChamber}{58.89}
\newcommand{\scalaSvmMeanIoUChamber}{78.66}

\definecolor{MyRed}{RGB}{0,0,0}
\definecolor{MyGreen}{RGB}{0,0,0}
\definecolor{MyBlue}{RGB}{0,0,0}

\newcommand{\TODOREV}[1]{\textcolor{black}{#1}}

\newcommand{\positiontextbox}[4][]{%
  \begin{tikzpicture}[remember picture,overlay]
    \node[inner sep=3pt,right,draw,line width=1pt,#1] at ($(current page.south west) + (#2,-#3)$) {\footnotesize{#4}};
  \end{tikzpicture}%
}

\newcommand{\DRAFT}[1]{\textcolor{black}{#1}}

\title{\LARGE \bf
Weather Influence and Classification with Automotive Lidar Sensors
}

\author{Robin Heinzler$^{1,3}$, Philipp Schindler$^{1}$, J{\"u}rgen Seekircher$^{1}$, Werner Ritter$^{2}$ and Wilhelm Stork$^{3}$
\thanks{$^{1}$Daimler AG, Benz-Str., 71063 Sindelfingen, Germany
        {\tt\small [firstname.lastname]@daimler.com}}%
\thanks{$^{2}$Daimler AG,  Wilhelm-Runge-Str. 11, 89081 Ulm, Germany
        {\tt\small werner.r.ritter@daimler.com}}%
\thanks{$^{3}$Institute for Information Processing Technology (ITIV), Karlsruhe Institute of Technology (KIT), Germany
		{\tt\small [firstname.lastname]@kit.edu}}%
}

\begin{document}
\maketitle
\thispagestyle{empty}
\pagestyle{empty}


\positiontextbox{1.505cm}{-1.4cm}{This paper has been submitted to the IEEE for publication. Copyright may be transferred without notice, after which this edition may no longer be available.}

\begin{abstract}
Lidar sensors are often used in mobile robots and autonomous vehicles to complement camera, radar and ultrasonic sensors for environment perception. Typically, perception algorithms are trained to only detect moving and static objects as well as ground estimation, but intentionally ignore weather effects to reduce false detections. In this work, we present an in-depth analysis of automotive lidar performance under harsh weather conditions, i.e. heavy rain and dense fog.
An extensive data set has been recorded for various fog and rain conditions\textcolor{black}{, which is the basis for the conducted in-depth analysis of the point cloud under changing environmental conditions.} In addition, we introduce a novel approach to detect and classify rain or fog with lidar sensors only and achieve an \textcolor{MyRed}{mean union over intersection of \vlpSvmMeanIoUChamber\,\% for a data set in controlled environments}. The analysis of weather influences on the performance of lidar sensors and the weather detection %
are important steps towards improving safety levels for autonomous driving in adverse weather conditions
by providing reliable information to adapt vehicle behavior.
\end{abstract}
%
\section{INTRODUCTION}
\begin{figure}[tb]
   	\centering
	 \includegraphics[width=1.0\linewidth]{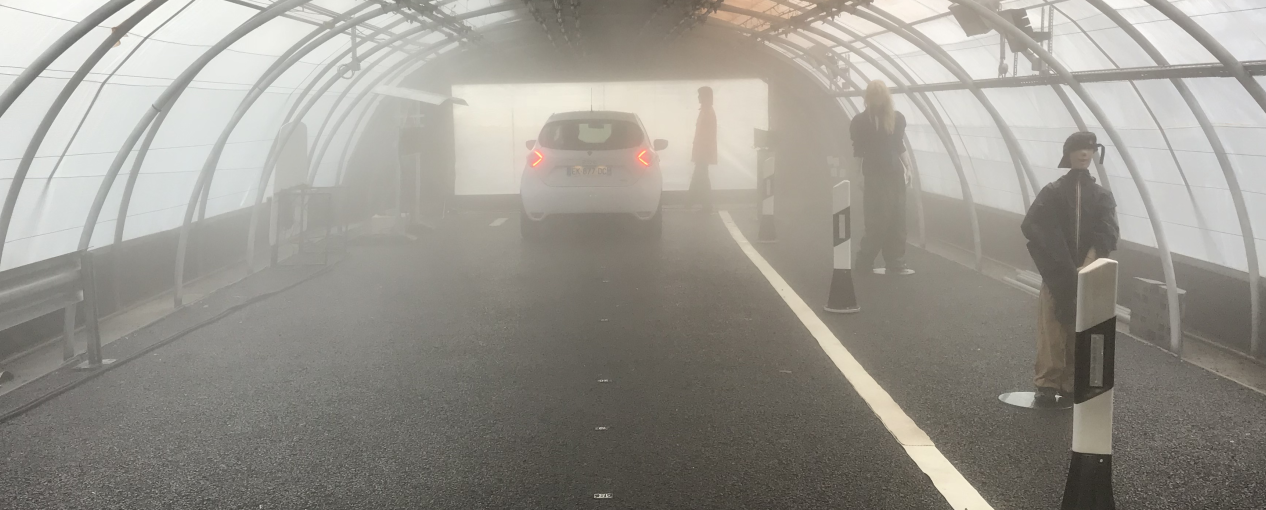}
	\caption{Experimental setup in the climate chamber in fog with a visibility of $20$-$30\,$m.}
	\label{fig:fog_example}
\end{figure}

Environment perception is a key challenge for autonomous driving and a major restricting factor for the availability and performance of the system. 
In order to increase both, the degree of automation and the availability of the system, various environmental conditions are to be considered, recognized by the system and reacted properly to.
In order to develop a robust perception and sensor fusion, it is of utmost importance to know potential degradation of different sensor types and mitigate their impact.
To develop a truly autonomous vehicle, it needs to recognize system boundaries without external intervention and react accordingly to master all environmental impairments. 
According to the definition of levels of automation by the 'Society of Automotive Engineers' (SAE), level 3 and 4 systems need the detection of system boundaries in order to hand over to a manual driver (3) or to bring the system into a safe state (4). For fully automation (level 5) mastery of all environmental conditions is required \cite{SAEInternational.2014}.
Therefore, the investigation of environmental influences on sensor performance is a crucial area of research for autonomous driving: sensors are facing ever changing and always unique weather conditions, such as dense fog in fig. \ref{fig:fog_example}.
For lidar sensors the most challenging environmental conditions are bright sun, fog, rain, dirt and spray.

In this paper we investigate the influence of rain and dense fog on state-of-the-art lidar sensors. The paper is structured as follows: Chapter 2 discusses related works and states the main contributions of this article, Chapter 3 introduces the details of the experimental setup, Chapter 4 describes the method of evaluating the recorded data set and Chapter 5 deals with the experimental results, the influence of rain and fog on lidar sensors and the detection of rain and fog based on lidar output data. Finally Chapter 6 draws a conclusion and gives an outlook to future work. 

\section{RELATED WORK}
Although, state-of-the-art data sets are commonly recorded at favourable weather conditions (e.g. \cite{AndreasGeiger.2013,GauravPandey.2011}), there exists a sizable amount of literature about the impact of harsh weather conditions such as fog, rain, dust or snow for lidar sensors \cite{AlNaboulsi.2004,Peynot.2009,Filgueira.2017,Hasirlioglu.2017,Hasirlioglu.2016,Hasirlioglu.2016b,Ijaz.2012,Rasshofer.2011,Ryde.2009,Sallis.2014,Shamsudin.2016, Phillips.2017,Bijelic2018Benchmark}:

\subsection{Lidar Sensors in Adverse Weather Conditions}
In 2009 a first data set with radar, visual camera, infrared camera and lidar sensors was recorded in challenging environment situations (dust, rain and smoke) by Peynot et al. \cite{Peynot.2009}. According to the results there is a significant attenuation for lidar sensors compared to radar sensors in challenging environment conditions. For lidar sensors it was also proven that objects could disappear behind the airborne dust. Based on the different attenuation of the two sensor concepts, an algorithm was developed to remove the dust reflections by filtering the laser data based on radar data. 

Hasirlioglu et al. proposed a theoretical model considering multiple reflections by rain drops or fog to determine the influence of fog and rain for automotive perception sensors \cite{Hasirlioglu.2016b}. The principle of the model is based on a longitudinal layer representation. Within each layer a reflection, transmission and absorption could occur. This approach considers multiple reflections and is verified with a developed fog and rain simulator \cite{Hasirlioglu.2017,Hasirlioglu.2016}. The developed system provides a maximum length of 4\,m and a overall distance between sensor and target of 10\,m. The influence of rain or fog is evaluated on a standardized Euro NCAP Vehicle Target (EVT) \cite{Sandner.2013}\textcolor{MyRed}{, which is optimized to represent a vehicle for visual camera sensors and not for lidar sensors. Therefore we have used real objects to ensure a correct reflection behavior for the backscattered light.} 
The results of the test setup shows that \textcolor{MyRed}{in general} radar sensors are robust against fog compared to lidar and camera sensors which are strongly affected by fog \cite{Hasirlioglu.2017}. 
\textcolor{MyRed}{As there is no quantification of the fog density (e.g. meteorological visibility), a direct conclusion on real environmental conditions is not possible.}
\DRAFT{
The lidar data set in dense fog by \cite{Bijelic2018Benchmark} enables this conclusion, as the data set was recorded in a dedicated climate chamber \cite{Colomb.2008} with a closed-loop controlled visibility range, as our data set. 
The detailed analysis takes the total number of scan points in a single frame, the intensity, the maximum detection distance and as sensor parameter tuning into account.
According to \cite{Bijelic2018Benchmark} the detection range of state of the art lidar sensors brakes down below $40\,$m visibility and is limited to $25\,$m, even with multiple returns.
}

\textcolor{MyRed}{The influence of rain for lidar sensors was analyzed in a similar fashion in \cite{Filgueira.2017,Ryde.2009}.} 
Filgueira et al. presented a work that quantifies the influence of rain for one lidar sensor \textcolor{MyRed}{and a static scene}; in detail the average range, intensity and number of points for certain objects \cite{Filgueira.2017}. The results show smaller changes in the distance of detected objects, while the intensity and the number of points decrease dramatically. 

\textcolor{MyRed}{In \cite{Ryde.2009} and \cite{Phillips.2017} the influence of dust is analyzed with one type of 2D laser scanner in detail, similar to our approach of studying the influence of fog and rain. Smoke and rain are additionally examined in \cite{Ryde.2009}, but since the utilized chamber couldn't produce artificial fog, there are no investigations about the influence of fog. According to \cite{Phillips.2017} the influence of dust on lidar sensors is systematic and predictable, as the lidar measures the leading edge of a dust cloud, which occurs with the lidar used from a transmission of about 70\,\%.}

Wojtanowski et al. \cite{Wojtanowski.2014} presented a very detailed inspection and discussion for range degradation of \textcolor{MyRed}{hypothetical} lidar sensors with 0.9\,$\mu$m and 1.5\,$\mu$m in fog and rain environmental conditions. 
Considering only the attenuation by fog, rain and wet surfaces, lidar sensors with a wavelength of 905\,nm are outperforming sensors with 1550\,nm. As air humidity did not influence the sensor's performance significantly, fog is the most suppressing factor \cite{Wojtanowski.2014}.
\DRAFT{Recently, Kutila et al. \cite{Kutila.2016,kutila2018automotive} analyzed the influence of harsh weather conditions for lidar sensors at $905\,\mathrm{nm}$ and $1550\,\mathrm{nm}$ by evaluating the signal to noise ratio (SNR) of the back-scattered light and the quantitative comparison of the number of points per object. According to \cite{kutila2018automotive} the $1550\,\mathrm{nm}$ lidar sensor is outperforming the $905\,\mathrm{nm}$ sensor in adverse weather, due to the lower restrictions on emitted light power to reach laser class $1$.
}

An approach for removing near field reflections caused by fog is proposed in \cite{Shamsudin.2016} \textcolor{MyRed}{for one static scene}. The algorithm is based on the light beam penetration, the intensity and geometrical features for measurements recorded in a fog-filled room with 2 and 6\,m visibility. \textcolor{MyRed}{Since the data sets used in \cite{Shamsudin.2016} only contains a single static scene, the reported results might be affected by overfitting of the underlying classifier.}

Lidar sensor performance significantly depends on the environmental conditions as demonstrated in \cite{AlNaboulsi.2004,Filgueira.2017,Ijaz.2012,Rasshofer.2011,Ryde.2009,Sallis.2014,Shamsudin.2016}. 
Thus, it is fundamental to recognize and quantify the impact of current weather on the lidar performance in order to develop robust perception and thus fusion algorithms. 
Foremost, for fusion and trajectory planning of autonomous cars, it is important to reliably classify current sensor performance for weighing sensor modalities optimally. As a consequence, the evaluation of the impact of various environmental conditions for a specific sensor system is essential.%
\subsection{Main Contribution}	
\textcolor{MyGreen}{The contributions of this paper are twofold:
\begin{itemize}
 \item First, we present a detailed analysis of the impact of rain and fog on different lidar sensors using a novel data set that excels in terms of realism, controllability, size and scenarios. 
 \item Second, we present a novel approach of detecting weather using lidar sensors only that obtains state-of-the-art performance.
\end{itemize} }

The recorded data set surpasses the spatial limitations of \cite{Hasirlioglu.2017,Hasirlioglu.2016,Ryde.2009} and additionally offers a closed loop controlled fog density by the meteorological visibility range from 20-60\,m and a stabilized rainfall rate at 55\,mm/h \cite{Colomb.2008}. Furthermore, in contrast to \cite{Hasirlioglu.2017,Hasirlioglu.2016,Ryde.2009} numerous dynamic scenarios with real traffic situations have been recorded under controlled weather conditions. \DRAFT{Compared to \cite{Bijelic2018Benchmark} and \cite{kutila2018automotive}, we used the same climate chamber, but a different sensor setup and data analysis strategy. While the analysis of \cite{Bijelic2018Benchmark} is focused on the maximum detection range, sensor parameter tuning and their influence on the point cloud, our approach is concentrating on the detection of adverse weather conditions and an analysis of the perception on object level. 
Moreover, our data set contains road recordings at sunny, cloudy and rainy weather conditions at day- and nighttime.}

Besides, determining the degradation of sensor performance, it is also significant to reliably detect and classify environmental conditions by the sensor itself. The idea to detect weather conditions with vehicle on-board sensors is stated in \cite{Zhu.2015}. In this paper, a novel approach of detecting weather conditions only by evaluating output data of a lidar sensor is developed. Thus the sensor determines its current performance and provides important information for any type of sensor fusion algorithm, in order to adapt the system behavior according to the current perception performance and environmental conditions (e.g. reduction of speed during precipitation and wet road). 
\DRAFT{
	In comparison to \cite{Shamsudin.2016}, it is not the goal to detect at point level whether the detection was caused by adverse weather or not. 
	Since disturbance points caused by fog or rain are not the main influencing factor, but rather the enormous reduction of sensor performance in terms of range. 
	Even if the interference points are filtered, the performance degradation remains and restricts the operation of the autonomous vehicle. Therefore, we use this information to identify sensor and system boundaries. 
	In this regard it is important to mention, that the input data of each sensor type is regarded separately in order to guarantee redundancy and avoid cross-dependencies. Thus, filtering raw data input of a specific sensor by another, as shown in \cite{Peynot.2009}, is not regarded in this paper. The information of the weather classification is therefore rather used to adjust confidence values of a fusion algorithm for different types of sensors separately. 
	In the long term, the goal is to enable both filtering of interference points caused by adverse weather and classification of system degradation.}		
\section{EXPERIMENTAL SETUP}
%
\begin{figure}[tb]
\includegraphics[width=1.0\linewidth]{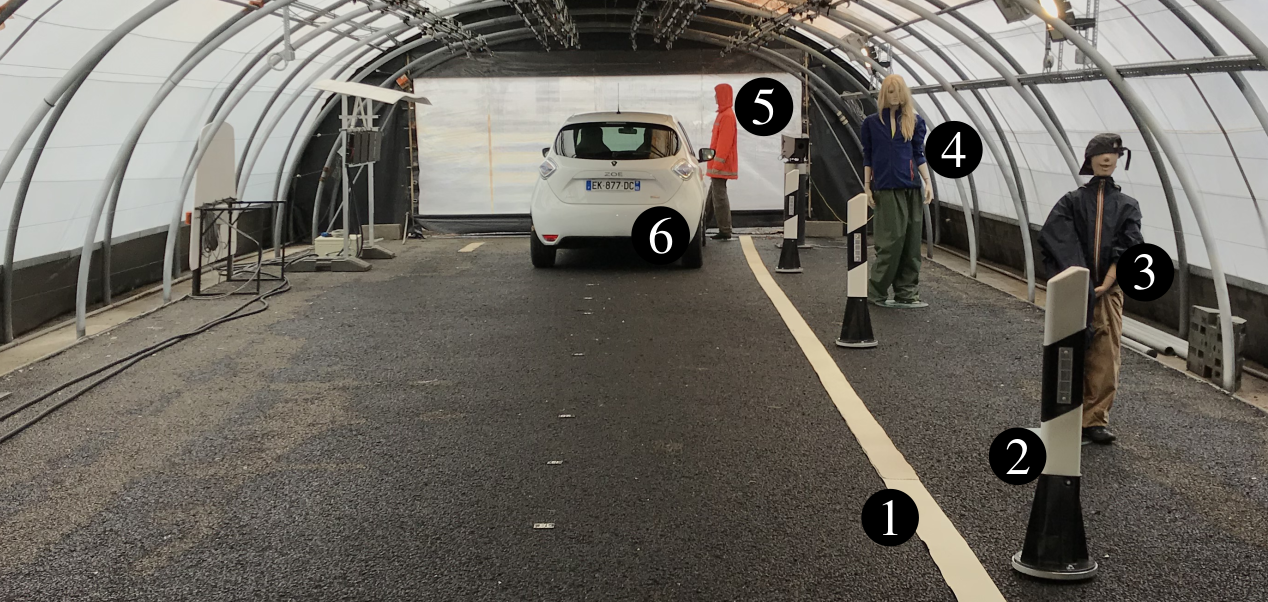}
	\caption{Experimental setup in the fog chamber for dynamic scenes in reference condition 'clear' without any rain or fog. Lane marking (1), reflector post (2), child and woman mannequin (3,4) and man mannequin with reflective vest (5) are stationary objects. Pedestrian, cyclist (both not shown) and car (6) are movable objects.}
	\label{fig:Setup_D2}
\end{figure}%
\begin{figure}[tb]
   	\centering
 \subfloat[national road, light rain\label{1a}]{
        \includegraphics[trim=1cm 4cm 1cm 1cm,clip,width=0.45\linewidth]{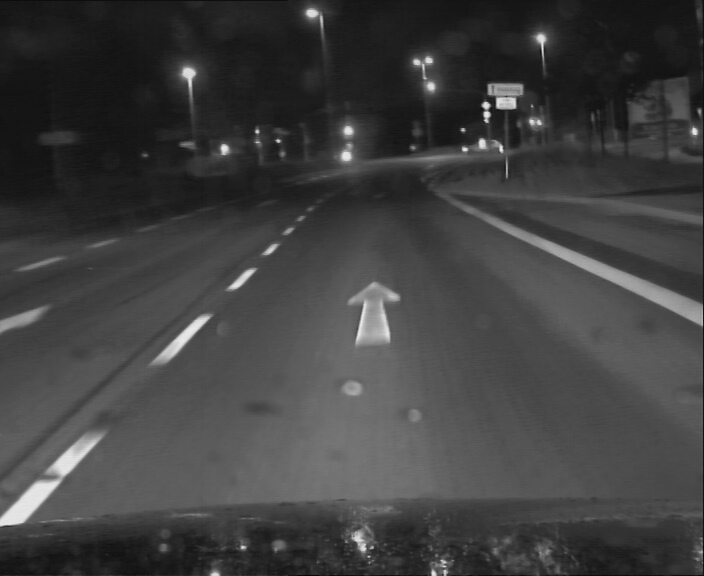}}
    \subfloat[rural road, heavy rain\label{1b}]{    
        \includegraphics[trim=1cm 4cm 1cm 1cm,clip,width=0.45\linewidth]{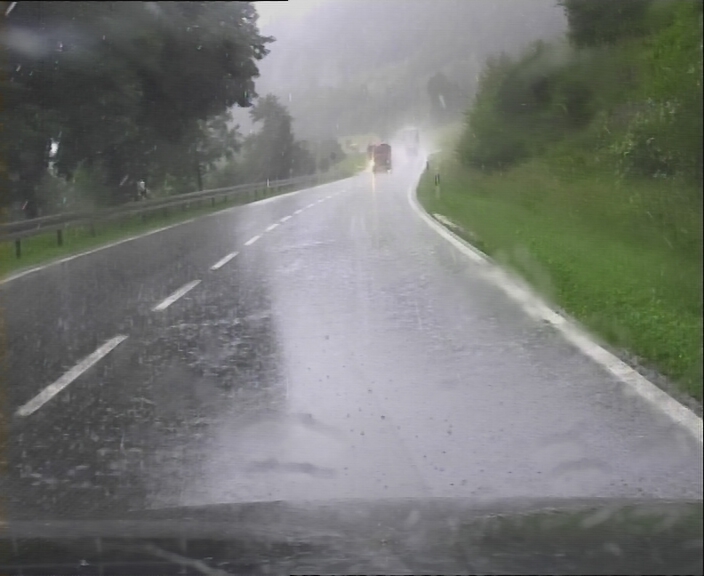}}\\
    \subfloat[roundabout in a city, no rain\label{1c}]{
        \includegraphics[trim=1cm 4cm 1cm 1cm,clip,width=0.45\linewidth]{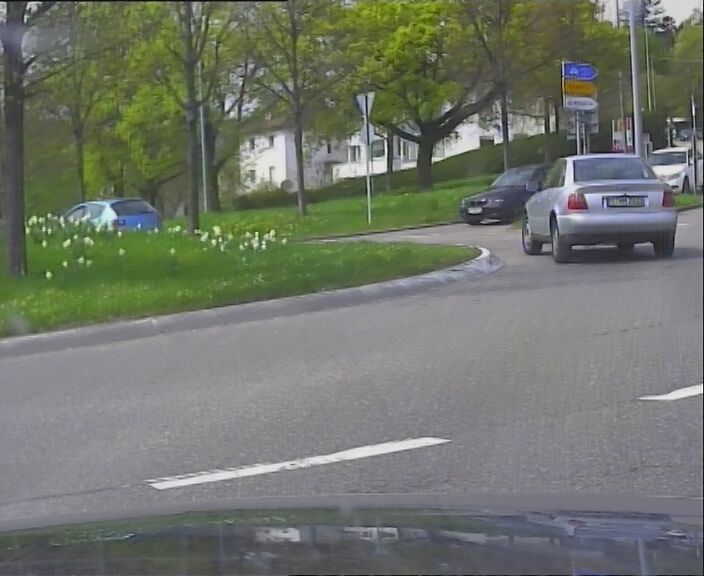}}
    \subfloat[traffic jam on a highway, no rain \label{1d}]{
        \includegraphics[trim=1cm 5.25cm 1cm 1cm,clip,width=0.45\linewidth]{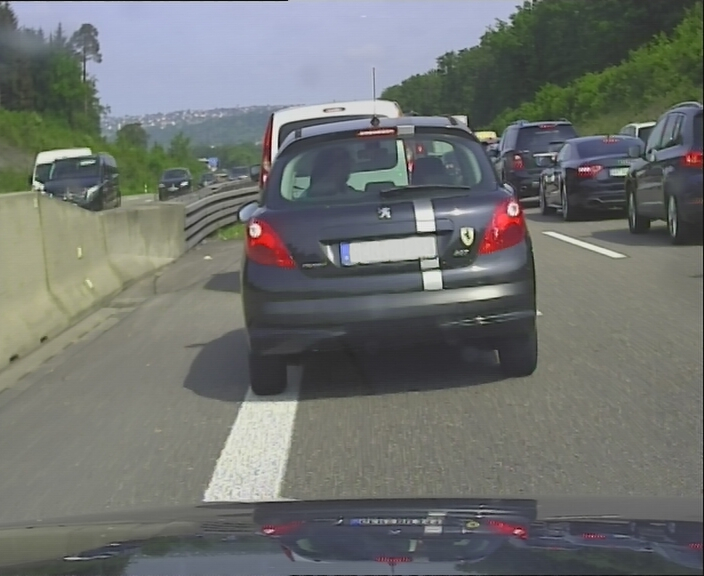}}
	\caption{\textcolor{MyRed}{
    Examples for the road data set under different environmental conditions, traffic situations and road types. All examples shown have been recorded during the day, with the exception of \ref{1a} recorded at night.}}
	\label{fig:road_vlp_data}
\end{figure}

Defined and reproducible weather conditions are essential to obtain a meaningful data set of perception sensors in different weather conditions. In Europe CEREMA's climate chamber in Clermont Ferrand (France) is the only 
publicly accessible climate chamber with the ability to produce homogeneous advection and radiational fog with a controlled and constant visual range \cite{Colomb.2008}. Additionally, the chamber provides a rain simulator with a stabilized rainfall rate. 

Based on the data set in the climate chamber a detailed evaluation of the sensor output data is conducted in clear, foggy and rainy conditions. 
Furthermore, a classification procedure is developed to obtain information about the weather condition from the laser scanner's point cloud. 
Three different setups are recorded with static and dynamic scenarios to reduce the time-correlation of the data set \textcolor{black}{and prevent an overfitting of trained classifiers. The static setup (Setup A) includes retro reflective and diffuse reflecting objects with 5\,\%, 50\,\% and 90\,\% reflectivity. The first dynamic setup (Setup B) represents typical scenes of real traffic situations \textcolor{MyRed}{(e.g. crossing pedestrians, cyclists at the roadside, approaching or leaving car, etc.)} with 8 different variants \textcolor{MyRed}{containing different trajectories, combinations and occlusions of the moving objects}, one of them is shown in fig. \ref{fig:Setup_D2}. The scenario in fig. \ref{fig:Setup_D2} shows a leaving car in the ego lane and several static objects, which typically occur in real-life traffic situations. 
The second dynamic setup (Setup C) contains diffuse reflecting targets leaving the field of view of the sensor. Consequently the influence of changing weather conditions and changing scenarios can be analyzed separately.}

\textcolor{MyRed}{In addition, road data has been recorded at several different environmental conditions. 
The road data set contains 5 scenarios without any rain (4 at daytime, 1 at nighttime), 3 with occasional rain, and 4 with almost permanent rain (3 at daytime, 1 at nighttime). Furthermore, the data set includes many different traffic situations, such as traffic jam or empty highway and different types of roads (highway, rural road, inner city).
Fig. \ref{fig:road_vlp_data} shows four different examples of the data set. The driven speed of the data was between 0 and 160\,km/h.}

The sensor setup contains two state-of-the-art lidar sensors: the Velodyne 'VLP16' and Valeo 'Scala'. Both sensors operate at about 905\,nm wavelength with a scanning system. The main difference is the mechanical design of the scanning pattern. 
While the 'VLP16' utilizes a rotator to spin transmitter and receiver, the 'Scala'  keeps the transmitter and receiver fixed and deflects sending and receiving light with a rotating mirror. Another difference is that the 'Scala' sensor detects the echo pulse width (epw) of the received light pulses, whereas the Velodyne sensor measures the intensity of the received pulses. \DRAFT{Both sensors are able to detect multiple returns, which are referred below as echoes. While the 'Scala' sensor provides three echoes ordered by distance, the 'VLP16' provides the last and the strongest echo. If the last and the strongest echo are identical, the second strongest echo is provided \cite{Valeo.2018,VelodyneLidarInc..2017}. In order to obtain a uniform denomination, for the 'VLP16' the strongest or second strongest echo is denoted as $1$ and the last return as $2$. If no multiple reflections are detected, there are no valid points for echo $2$ or $3$ on the 'Scala' sensor, whereas for the 'VLP16' echo $1$ and $2$ are identical.}%
The recorded data set in the climate chamber contains about 274,000 frames for the 'VLP16' and 105,000 for the 'Scala' Sensor. The road data was recorded with the 'VLP16' and includes 270,000 frames.\\ %
\section{METHOD} %
\TODOREV{We compose a feature vector which reflects the impact of fog and rain on the raw output data of lidar sensors without preprocessed filtering. }
The point cloud at time $k$ is represented as matrix $P\in\mathbb{R}^2$ where each row $i=(1,\dots, n)$ contains one point with $j=(1,\dots,m)$ number of attributes. 
\begin{equation}
  	 P(k)^{n\times m} = 
   	\begin{pmatrix}
        p_{11}(k) & p_{12}(k) & \cdots & p_{1m}(k) \\
        p_{21}(k) & p_{22}(k) & \cdots & p_{2m}(k) \\
        \vdots & \vdots &        & \vdots \\
        p_{n1}(k) & p_{n2}(k) & \cdots & p_{nm}(k) \\
    \end{pmatrix} \; .
\end{equation}
The number and content of attributes is sensor specific but the same for each and every point. 
To ensure a universal structure the unification over all sensor specific signals is represented in 
\begin{equation}
 p_{ij} = (p_{i1}, \dots, p_{ij}, \dots, p_{im})
 \label{eq:p_ij}
\end{equation}
where each column represents a single attribute.
In order to ensure a clear notation, the column index of a point cloud is omitted, but marked with a variable per column.
\begin{equation}
    p_{i} = (x_{i}, y_{i}, z_{i}, r_{i}, \theta_{i}, \varphi_{i}, e_{i},I_{i}, epw_{i})\; .
    \label{eq:p_i} 
\end{equation} 
The corresponding notation is: ($x,y,z$) for the cartesian and ($r, \theta, \varphi$) for the spherical coordinates, $e$ for the echo number, $I$ for the intensity and $epw$ for the echo pulse width.  

As the return energy of light scattered by atmospheric particles is weak, impact of ambient conditions are mainly expected at close range. Thus the point cloud is spatially filtered, restricting processing to the near-range \textcolor{black}{($x\leq20m$)} of the ego-lane \textcolor{black}{($-1.5m\leq y\leq+1.5m$)}, for the following analysis of the influence of fog and rain on the point cloud attributes.
\textcolor{MyRed}{The  focus on a region of interest (ROI)} reduces dependencies of the composition of the point cloud on \textcolor{MyRed}{a specific scenario (e.g. guardrails or vegetation at the roadside) and saves computation time.}
  
A distinction by the echo number $t\in\mathbb{N}$ representing the first, second or third return signal of a transmitted light pulse, is reasonable as the number of echoes per scan direction relates 
not only to the number of objects but also atmospheric particles which are potential scatter points for the light.
Hence, the amount $M_t$ is defined as $M_t:=\{e_i|e_i = t\}$ with $t\in\{1,2,3,\cdots\}$ being the number of the respective received return pulse per angle. For the number of points for a specific echo, the signal $N_t(k)$ is derived:
\begin{equation}
	N_t(k) = |e_i(k)| \quad \forall e_i\in M_t\; .
\end{equation}  	
The mean and variance of one attribute $p_j$ is calculated for each frame by:
\begin{equation}
	\overline{p_j}(k) = \frac{1}{n} \sum_{i=1}^{n} p_{ij}(k) \quad \mathrm{var}(p_j(k))= \frac{1}{n} \sum_{i=1}^n (p_{ij} - \overline{p_{ij}}(k))^2 \; .
\end{equation}
For example the mean distance of all points corresponding to a specific return is given by:
\begin{equation}
		\overline{r}_t(k) = \overline{r_{i}}(k) \quad \forall e_i\in M_t\; .
\end{equation}

The spatial distribution of the points is represented by the eigenvalues of the covariance matrices of $x$, $y$ and $z$, similar to \cite{Shamsudin.2016}.

\textcolor{black}{Finally, the assignment of the resulting feature vector $f=(f_1, \dots, f_{16})^T$, shown in table \ref{table:feature_vector}, describes one frame of the laser scanner.
The features are down selected by a neighboring component analysis to find \textcolor{MyRed}{
the parameters with the highest effect \cite{yang.2012}}.}
Additionally, the impact of the rather static scene has to be mitigated to not bias the training of ambient condition detection.
For example, the total number of points is not taken into consideration for weather classification because it is highly dependent on the scenario (empty highway vs. inner-city traffic jam).  
\begin{table}[h] 
	\centering
	\caption{Feature vector for environment classification based on point cloud data. For the feature set of 'VLP16' the echo pulse width $epw$ is replaced by the intensity $I$.}
	\label{table:feature_vector}
	\begin{tabular}{p{0.15cm}p{0.5cm}|p{0.15cm}p{1.0cm}|p{0.15cm}p{1.1cm}|p{0.15cm}p{1.3cm}}
		\toprule
		$f_1$ 	& $N_1$   		        &$f_5$ 		& $\overline{r_2}$      &$f_{9}$	& $\mathrm{mean}(r)$       &$f_{13}$ 	& $\mathrm{mean}(epw)$	\\	
		$f_2$	& $N_2$					&$f_6$ 		& $\overline{r_3}$		&$f_{10}$ 	& $\mathrm{mean}(\varphi)$ &$f_{14}$ 	& $\mathrm{eig}(\mathrm{cov}(x))$\\
		$f_3$	& $N_3$                 &$f_7$ 		& $\mathrm{mean}(e)$    &$f_{11}$ 	& $\mathrm{mean}(\theta)$  &$f_{15}$ 	& $\mathrm{eig}(\mathrm{cov}(y))$\\
     $f_4$ 	& $\overline{r_1}$ 		&$f_8$ 		& $\mathrm{var}(e)$	    &$f_{12}$ 	& $\mathrm{var}(epw)$      &$f_{16}$ 	& $\mathrm{eig}(\mathrm{cov}(z))$ \\		
		\bottomrule
	\end{tabular}
\end{table} %
\section{EXPERIMENTAL RESULTS} %
\begin{figure}[tb]
	\centering
	\def\svgwidth{1.0\linewidth}
  {\footnotesize\import{img/fig_20171205_scala_layer-2_point_cloud_echo_image/}{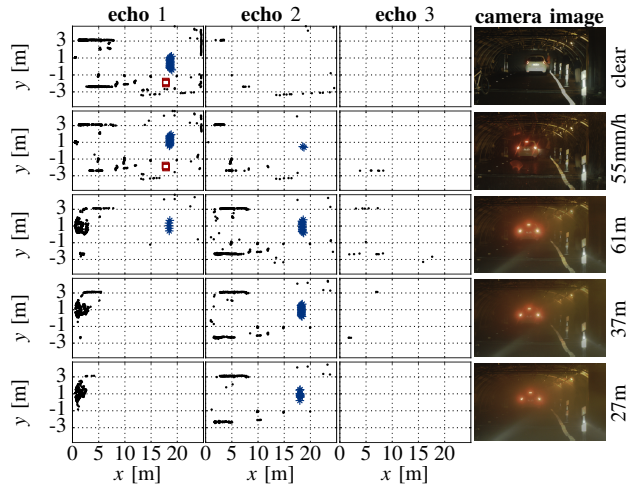}}
	\caption{Point cloud of 'Scala' sensor in birds-eye view for a single \textcolor{black}{random but representative} frame $k$, split by echo number. \textcolor{MyRed}{The fig. visualizes} the scene of Setup B in the chamber and the resulting point cloud. \DRAFT{An image of a sate-of-the-art automotive camera is given at the right side, while the meteorological visibility $V$ is stated in meter rightmost in case of fog and the rainfall rate in $\mathrm{mm/h}$. For reference conditions the label 'clear' is given. The pedestrian mannequin is highlighted with red boxes (object no. 4 in fig. \ref{fig:Setup_D2}).} and the car in blue stars (object no. 6 in fig. \ref{fig:Setup_D2}). All other points are marked as black dots.}
	\label{fig:echo_scala1_pointcloud}
\end{figure}
First, we discuss \DRAFT{the influence of different weather conditions on point clouds and object perception, second we discuss the previously determined features and
third a classification procedure is presented to extract the weather condition from the laser scanner's point cloud.} The ground truth measurements are the meteorological visibility $V$ in $\mathrm{m}$ and the rainfall rate $R$ in $\mathrm{mm/h}$ and are provided by the climate chamber\textcolor{MyRed}{\cite{Colomb.2008}}. \textcolor{black}{For road recordings the ground truth is given by the signal of the so called rain light sensor, which measures the rain intensity to automatically adjust the speed of the wiper.}
\DRAFT{\subsection{Weather Influence on Point Clouds and Object Perception}}
\begin{figure}[tb] %
	\centering %
	\def\svgwidth{1.0\linewidth} %
	{\footnotesize\import{img/fig_20171205_scala_layer-2_boxplot/}{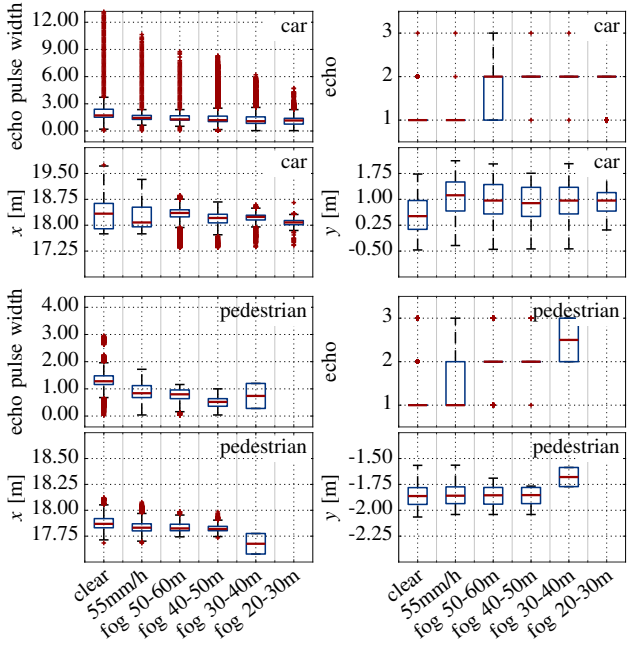}} %
	\caption{\DRAFT{Object perception for \textit{'Scala'} sensor with $1,200$ frames per weather condition (except for rain, $921$ frames). The boxplot shows the result of the corresponding raw point cloud ($x$- and $y$-coordinate, echo and epw) for a \textit{car} and \textit{pedestrian}. The weather is shown on the ordinate axes ordered by descending meteorological visibility $V$. In case of fog the visibility is stated in $\mathrm{m}$, for rain the rainfall rate in $\mathrm{mm/h}$ and the label 'clear' for reference conditions. The locations of the target objects are given in fig. \ref{fig:echo_scala1_pointcloud}}} %
	\label{fig:scala1_boxplot_object} %
\end{figure} %
\begin{figure}[tb] %
	\centering %
	\def\svgwidth{1.0\linewidth} %
	{\footnotesize\import{img/fig_20171205_vlp16_layer-0-1-2-3-4-5-6-7-8-9-10-11_boxplot/}{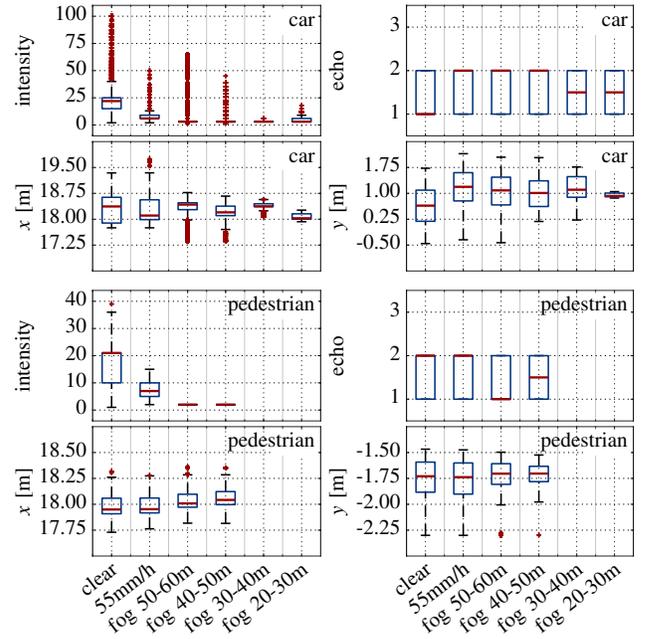}} %
	\caption{\DRAFT{Object perception for \textit{'VLP16'} sensor with $1,200$ frames per weather condition. The structure of this fig. is identical to fig. \ref{fig:scala1_boxplot_object}.}} %
	\label{fig:vlp16_boxplot_object} %
\end{figure} %
\DRAFT{Fig. \ref{fig:echo_scala1_pointcloud} shows the point cloud of 'Scala' sensor in birds-eye view for a single random but representative frame $k$, split by echo number.
Additionally, an image of a state-of-the-art automotive camera sensor is given rightmost. The depicted point cloud is taken from setup B shown in fig. \ref{fig:Setup_D2} and illustrates the scene and the positions of the objects.
For a statistical comparison of the object perception performance, the accumulation of points which are corresponding to a car or pedestrian object of at least $1,200\,$ frames for each weather condition are shown in a boxplot in fig. \ref{fig:scala1_boxplot_object} and \ref{fig:vlp16_boxplot_object}.}

The point cloud with strong rain (55\,mm/h) shows less points at the end of the climate chamber compared to the point cloud without any simulated weather. This relates to a reduced detection range of the sensor.
The detection quality of objects such as cars is highly important and interesting:
\DRAFT{For example the car positioned in about $19\,$m distance (highlighted in blue) is detected by both sensors in all scenarios, as shown in fig. \ref{fig:scala1_boxplot_object} and \ref{fig:vlp16_boxplot_object}. In clear conditions always the first return is received from the car, while in fog and rain the second echo contributes the majority to the detection of this car.}
Consequently, the occurrence of second echoes on objects can be an indication for the presence of fog or rain.

In fog, at a visibility range from $50$-$60$m, a large number of first echoes is observed at very short distance. Moreover, the detection quality and range is expected to be impaired as significant laser power is scattered by the atmospheric particles, leading to the other echoes. The environment perception and the range of the sensor is limited.
Only few secondary echoes can be associated with the fog as most coincide with the position of the car.
In dense fog (visibility at $20$-$40$m) the environment perception is highly limited.
Nearly all primary echoes are observed at a range of less than 5 m and thus caused by the fog.
Nevertheless highly reflecting targets like the retro reflectors of the tail lights are still correlated with secondary or tertiary echoes.
Comparing all fog and rain measurements with the clear ones, the number of second and third returns increases (fig. \ref{fig:scala1_boxplot_object} and \ref{fig:vlp16_boxplot_object}).
Additionally, our evaluation shows, that a multi-echo sensor is beneficial as it returns also weaker reflections such as fog and rain, while maintaining reasonable object detection performance compared to single return sensors.

\textcolor{MyRed}{Comparing these results with the four range measurement behaviors of lidar sensors in the presence of dust, introduced in \cite{Phillips.2017}, the influence of fog is similar to dust, in which the measuring range is the front of the dust cloud. Whereas the influence of rain seems to be different.} %

\DRAFT{
Mainly first echoes are received for rain and clear conditions in fig. \ref{fig:scala1_boxplot_object} and \ref{fig:vlp16_boxplot_object}. Only few second and third echoes are provided during rain.
The variance of the epw is continuously decreasing with lowering visibility range. This observation holds true for the intensity measured by 'VLP16'.
Consequently, object detection algorithms that leverage intensity or epw information are likely to be strongly influenced by adverse weather conditions. }
In addition, with an decreasing visibility range, the measured distance of the car decreases slightly as well as the number of outliers in terms of distance accuracy. Furthermore, the majority of the received points are detected as second returns.

\begin{figure}[tb]
	\centering %
	\def\svgwidth{1.0\linewidth} %
	{\footnotesize\import{img/fig_20171205_scala_vlp16_point_density/}{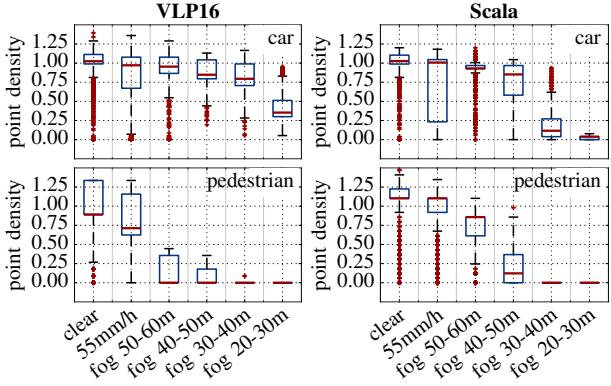}} %
	\caption{\DRAFT{Object point density for a \textit{car} and \textit{pedestrian} object. The density is the number of points per object in one frame scaled by the average number of points per object in clear conditions.}} %
	\label{fig:scala1_vlp16_obj_point_density} %
\end{figure}

\DRAFT{
Finally, in order to access the influence on the perception performance, a point density is calculated as a key metric, quantifying the impact of missing points.
The density rate is based on the total number of points $N_{t}^O(k)$ from object $O$ in frame $k$ and scaled by the mean over all frames in reference conditions without any fog or precipitation. Hence, the object density is an indication of the degradation of the object perception. In fig. \ref{fig:scala1_vlp16_obj_point_density} the resulting density is illustrated for a pedestrian and a car for both sensors. 
As a result, the perception of the car at $19\,$m remains rather robust during rainfall with $55\,$mm/h and degrades in fog with a visibility range of $20-30\,$m to a median of $0.36$ for 'VLP16' and even to $0.04$ for 'Scala'. In contrast, the detection density for the pedestrians at approximately $18\,$m declines significantly to $0.72$ in rain for 'VLP16' and remains rather robust for the 'Scala' sensor. In fog with a visibility range between $50-60\,$m the pedestrian is mostly not detected by the 'VLP16'. The 'Scala' sensor is able to detect the pedestrian with a density of $0.87$ down to a visibility of $50-60\,$m. 
Below a visibility of $40\,$m the detection density for the pedestrian is $0$. 
Consequently, objects without any retro-reflective materials are not perceived by lidar sensors in dense fog, even at close range.}
\DRAFT{\subsection{Weather Influence on Feature Vector}}
\begin{figure}[tb]
  	\centering
	\def\svgwidth{1.0\linewidth}
  {\footnotesize\import{img/fig_feat_discardedxy_mean_intensity_Day1_150300_183243_vlp_scala_boxplot/}{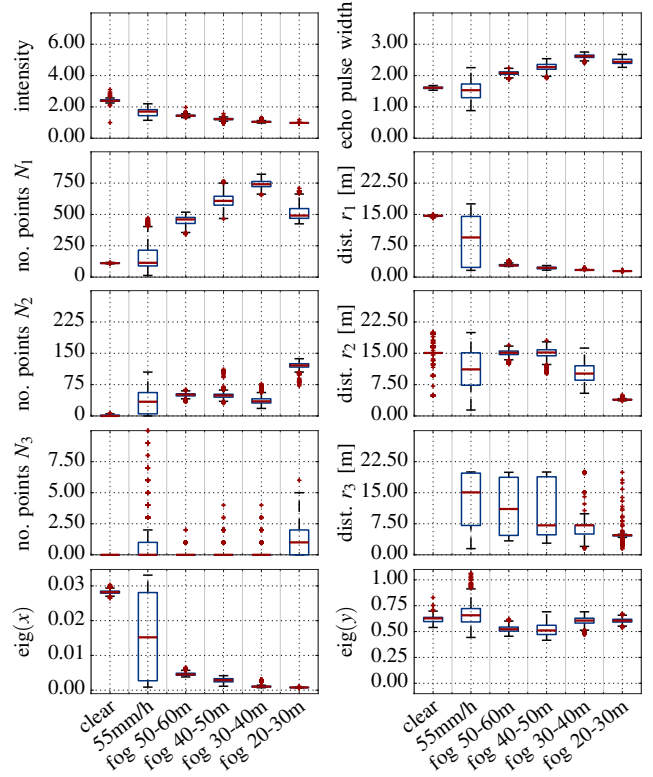}}
	\caption{
		Analysis of environmental influences on lidar point clouds. All measurements represent the same \textit{static} scene to investigate the influence of the weather only. Each column within a subfigure denotes one distinct weather condition \textcolor{MyRed}{with at least $1,200$ frames}, ordered by descending visibility.
		The stated mean intensity ($f_{13}$) is taken from 'VLP16' lidar sensor. All other signals are based on 'Scala' measurements, as 'VLP16' behavior is comparable. With the difference that the 'VLP16' only outputs two echoes, which yields differences in the number of points for second and third echo.
		The intensity $f_{13}$, epw $f_{13}$ and the distances $\overline{r}_t$ ($f_{4,5,6}$) are the mean values over all points of one frame. The number of points $N_t$ ($f_{1,2,3}$) and mean distances $\overline{r}_t$ are derived for each echo $t\in{1,2,3}$ for the first, second or last return separately. The eigenvalues ($f_{13,14,16}$) were calculated from the covariance matrix of all points.
	}
	\label{fig:mean_epw_intensity_n_r_D1}
\end{figure}
In fig. \ref{fig:mean_epw_intensity_n_r_D1} selected features are illustrated for more than $1,200$ frames per weather condition of the static setup A.
Considering the number of points for each return $N_t(k)$, 
it is to be expected that the number of second and third returns will increase with the presence of fog and rain due to multiple reflections. In fig. \ref{fig:mean_epw_intensity_n_r_D1} the different weather conditions are discernible in the signal $N_{1,2}(k)$. There is a significant difference for the variance of the second echo $N_2$ in foggy, rainy and clear conditions. It is also interesting, that there is no distinctively difference in the number of $N_3$ for dense fog and rain.

The mean distance $\overline{r}_t$, which is calculated for each echo separately, seems to be a good measure for estimating the presence of fog or rain. The described signal is illustrated in fig. \ref{fig:mean_epw_intensity_n_r_D1} and shows a strong decrease for $\overline{r}_1$ in fog. At the same time, the variance is clearly greatest in the rain. The mean and variance of ${r}_{2,3}$ increases in rain and fog compared to clear conditions.

The paradox of increased \textit{epw} and reduced \textit{intensity} by fog is discussed in \cite{Gotzig.2014}.
Fig. \ref{fig:mean_epw_intensity_n_r_D1} confirms the claim of \cite{Gotzig.2014} as the epw from the Scala sensor increases in foggy conditions and is approximately inversely proportional to the fog density. Furthermore, reflections from rain drops show an smaller epw, as water drops in rain are less dispersed than in fog. As a result, the epw is highly influenced by weather and could be used as a signal to gain information about the local environmental conditions.
Regarding the intensity of the Velodyne sensor, there is only a small decrease of intensity in dense fog and the greatest variance in rain.

Furthermore, rain or fog are influencing the \textit{eigenvalues} of the covariance matrices of $x$ and $y$ ($f_{14,15}$).
While the presence of fog and rain is influencing the eigenvalue $\mathrm{cov}(x)$, no dependency on $\mathrm{eig}(\mathrm{cov}(y))$ can be derived.
This could be based on the symmetrical structure of the setup related to the y-axis. Due to the small field of view in $z$-direction, the eig($z$) is not evaluated.
In summary, the influence of rain and fog is visible in static scenes using the lidar point cloud.

\begin{figure}[tb]
  	\centering
  	\def\svgwidth{1.0\linewidth}
  {\footnotesize\import{img/fig_feat_discardedxy_mean_intensity_Day2_160000_193000_vlp_scala_boxplot/}{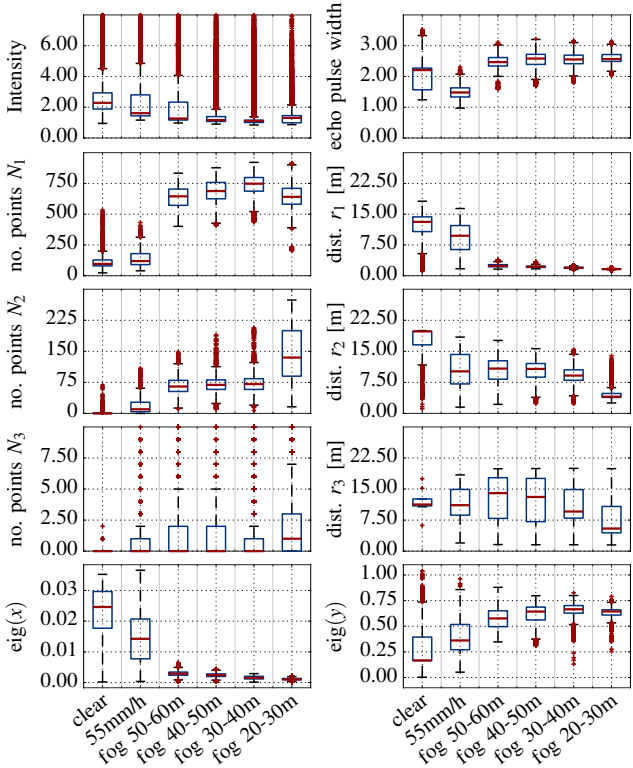}}
  	\caption{
		Analysis of environmental influences on lidar point clouds. All measurements represent the same \textit{dynamic} scene to investigate the influence of the weather and dynamic environments. The structure of this fig. is identical to fig. \ref{fig:mean_epw_intensity_n_r_D1}.}
  	\label{fig:mean_epw_intensity_n_r_D2}
\end{figure}

Next, we evaluate dynamic scenarios with the same methods. A setup of dynamic scenarios is repeated for the well-controlled environmental conditions.
The dynamic scenes mimic an approaching car, crossing cyclists and a pedestrian walking on the sidewalk.
Comparing the static scenes (fig. \ref{fig:mean_epw_intensity_n_r_D1}) with dynamic ones (fig. \ref{fig:mean_epw_intensity_n_r_D2}) the variance increases for all derived signals, while the difference of the mean decreases.
In addition the number of outliers increases significantly, especially for the intensity of the 'VLP16' laser scanner. This can be explained with the scenario of the approaching vehicle, since in this scenario for some frames retro reflective objects were in the immediate vicinity of the sensor. In conclusion, the pattern recognition task to cluster the different environmental conditions is more challenging in dynamic than in static scenarios.
\\
\subsection{Weather Classification by Means of Lidar Sensors}
For the development of a weather detection algorithm a $k$ Nearest Neighbor classifier (kNN) with $k$ = 10 and a Support Vector Machine (SVM) are applied.
The prediction feature vector is given in table \ref{table:feature_vector}. The response of the classifier is set to 'clear', 'fog' or 'rain'. The different visibility ranges in fog have not been taken into account for the classifier response, since the features do not differ very much in these conditions. 
\setlength{\tabcolsep}{3pt}
\begin{table}[h]
  \centering
  \caption{\textcolor{MyRed}{%
  The overall classification testing results for climate chamber and road data. The number of samples used for testing is stated in each row per class. As classification measures the true positive rate (TPR), false positive rate (FPR) and the intersection over union (IoU) are given. The classes are numbered as follows: 1 clear, 2 rain and 3 fog. The classifiers with the greatest performance in therms of mean IoU are printed in bold.}}
  \label{table:overall_results}
 
  \begin{tabular}{lllllllllll}
  \toprule
    \multirow{2}{*}{\rot{place}}&\multirow{2}{*}{\rot{clf}} & \multirow{2}{*}{\rot{class}}
    &\multicolumn{2}{c}{\# samples}&\multicolumn{2}{c}{TPR [$\%$]}&\multicolumn{2}{c}{FPR [$\%$]} &\multicolumn{2}{c}{IoU [$\%$]}\\
    & & &VLP&Scala&VLP&Scala&VLP&Scala&VLP&Scala\\\midrule
    \multirow{6}{*}{\rot{climate chamber}}& \multirow{3}{*}{\rot{kNN}}
    & 1 &5,558& 5,643& 93.91& 66.47& 6.09& 33.53& 93.85& 41.37\\
    && 2 & 10,566& 14,115& 97.52& 64.13& 2.48& 35.87& 95.86& 43.68\\
    && 3 & 92,708& 101,707& 99.98& 94.43& 0.02& 5.57& 99.48& 91.61\\
    & \multirow{3}{*}{\rot{SVM}}
    & 1 &5,558& 5,643& 100.00& 83.19& 0.00& 16.81& \textbf{96.29}& \textbf{53.34}\\
    && 2 & 10,566& 14,115& 95.86& 84.92& 4.14& 15.08& \textbf{95.78}& \textbf{83.70}\\
    && 3 & 92,708& 101,707& 99.80& 99.78& 0.20& 0.22& \textbf{99.35}& \textbf{98.95}\\\midrule
    \multirow{4}{*}{\rot{road}}&\multirow{2}{*}{\rot{kNN}}
    & 1 &33,369& \multicolumn{1}{c}{--}& 97.60& \multicolumn{1}{c}{--}& 2.40& \multicolumn{1}{c}{--}& \textbf{96.72}& \multicolumn{1}{c}{--}\\
    && 2 & 4,570& \multicolumn{1}{c}{--}& 92.45& \multicolumn{1}{c}{--}& 7.55& \multicolumn{1}{c}{--}& \textbf{77.04}& \multicolumn{1}{c}{--}\\
    & \multirow{2}{*}{\rot{SVM}}
    & 1 &33,369& \multicolumn{1}{c}{--}& 97.34& \multicolumn{1}{c}{--}& 2.66& \multicolumn{1}{c}{--}& 96.47& \multicolumn{1}{c}{--}\\
    && 2 & 4,570& \multicolumn{1}{c}{--}& 92.25& \multicolumn{1}{c}{--}& 7.75& \multicolumn{1}{c}{--}& 75.17& \multicolumn{1}{c}{--}\\
  \bottomrule
  \end{tabular}

\end{table}
\textcolor{MyRed}{
The different setups in the chamber are used to reduce the time correlation of the data set. Thus Setup A and B are used for training, while setup C is used for testing.
The mean union over intersection (IoU) for the 'VLP16' is \vlpKnnMeanIoUChamber\,\% (kNN) and \vlpSvmMeanIoUChamber\,\% (SVM) and thus exceptionally satisfactory. The classification result for the 'Scala' sensor is \scalaKnnMeanIoUChamber\,\% for the kNN and \scalaSvmMeanIoUChamber\,\% for the SVM classifier and thus significantly lower than the results of the 'VLP16', which could be caused by the significantly smaller vertical field of view and thus number of points per frame. Since the number of samples per class is not evenly distributed, the accuracy is not used to evaluate the classifiers in detail as illustrated in table \ref{table:overall_results}. Regarding the IoU per class, the kNN approach provides obviously not satisfying classification results for the classes 'clear' and 'rain' for the Scala sensor. The SVM achieves slightly better results for the class 'clear' and significantly better results for the class 'rain'.}

Since the weather conditions of the real-world and climate chamber data differ distinctly, they are considered separately. The splitting for training and verification for the road data is done similarly. Thus 4 recordings at clear conditions, 3 with occasional rain and 2 with permanent rain are used for training, while the remaining recordings are preserved for testing (1 recording with clear conditions, 2 with rain with night- and daytime). The subdivision of the data set is chosen in such a way that each data set has samples from every traffic scenario \textcolor{MyRed}{(empty road, traffic jam, inner city, etc.)} and every weather condition and at the same time a subdivision of $80\,\%$ to $20\,\%$ is given between training and testing. Thus time series effects can be avoided. The achieved mean IoU for the 'VLP16' is \vlpKnnMeanIoURoad\,\% with the kNN classifier. The IoU for the class rain is at $77.04\,\%$ significantly lower than the IoU of the class 'clear' ($96.72\,\%$). \TODOREV{The decrease of the IoU for rain in real-world environments could be caused by the larger variety of the rainfall rate and the lower accuracy of ground truth. } 
\section{CONCLUSION}
We presented an in-depth analysis of the influence of fog and rain on lidar sensors and introduced a novel approach to classify the weather status based on a laser scanner's point cloud for both controlled and uncontrolled environments. 
\DRAFT{The detailed analysis of object perception shows a significant reduction of the number of points per object and decreased variance of the measured intensity or epw values. 
As a result, the perception of objects is expected to be significantly impaired by adverse weather in addition to the reduction of detection range. Moreover, we expect an increase of misclassifications and even wrong detections due to the reduced contrast in intensity.
Thus, the recognition of these weather conditions becomes indispensable, especially for low reflective objects. }
The proposed \DRAFT{classification} approach achieves very satisfactory results for the majority of the classes. In order to reduces dependencies of the composition of the point cloud on a specific  scenario, a region of interest for weather classification is introduced. This also reduces the processing time, especially for high-resolution lidar sensors.

Compared to \cite{Hasirlioglu.2017,Hasirlioglu.2016,Shamsudin.2016}, we used a more extensive data with a larger range of meteorological visibility, stabilized with a closed-loop control, a larger spatial measuring range and rain data recorded in both controlled and uncontrolled environments.

Further extensions of our work can be achieved by applying advanced classification methods, an accumulation over time for the classifier result and a finer division of classes.  	
In addition, the ROI can be selected dynamically for example based on map data.
\section*{ACKNOWLEDGMENT}
We would like to express utmost appreciation to Florian Piewak, Mario Bijelic and Tobias Gruber for their valuable and constructive comments. Furthermore we would also like to thank Tatjana Immel and Ronald Singer for their support in evaluating the measurement data. The authors thank the DENSE project and CEREMA for the test facility. 
%

\bibliographystyle{IEEEtran}
\bibliography{rh}

\begin{thebibliography}{10}
\providecommand{\url}[1]{#1}
\csname url@samestyle\endcsname
\providecommand{\newblock}{\relax}
\providecommand{\bibinfo}[2]{#2}
\providecommand{\BIBentrySTDinterwordspacing}{\spaceskip=0pt\relax}
\providecommand{\BIBentryALTinterwordstretchfactor}{4}
\providecommand{\BIBentryALTinterwordspacing}{\spaceskip=\fontdimen2\font plus
\BIBentryALTinterwordstretchfactor\fontdimen3\font minus
  \fontdimen4\font\relax}
\providecommand{\BIBforeignlanguage}[2]{{%
\expandafter\ifx\csname l@#1\endcsname\relax
\typeout{** WARNING: IEEEtran.bst: No hyphenation pattern has been}%
\typeout{** loaded for the language `#1'. Using the pattern for}%
\typeout{** the default language instead.}%
\else
\language=\csname l@#1\endcsname
\fi
#2}}
\providecommand{\BIBdecl}{\relax}
\BIBdecl

\bibitem{SAEInternational.2014}
{SAE International}, ``{Taxonomy and Definitions for Terms Related to On-Road
  Motor Vehicle Automated Driving Systems},'' 2014.

\bibitem{AndreasGeiger.2013}
A.~Geiger, P.~Lenz, C.~Stiller, and R.~Urtasun, ``{Vision meets Robotics: The
  KITTI Dataset},'' \emph{International Journal of Robotics Research}, 2013.

\bibitem{GauravPandey.2011}
P.~Gaurav, R.~M. James, and M.~E. Ryan, ``{Ford campus vision and lidar data
  set},'' \emph{International Journal of Robotics Research}, pp. 1543--1552,
  2011.

\bibitem{AlNaboulsi.2004}
M.~{Al Naboulsi}, ``{Fog attenuation prediction for optical and infrared
  waves},'' \emph{{Optical Engineering}}, vol.~43, no.~2, p. 319, 2004.

\bibitem{Peynot.2009}
T.~Peynot, J.~Underwood, and S.~Scheding, ``{Towards reliable perception for
  unmanned ground vehicles in challenging conditions},'' \emph{IEEE/RSJ Int.
  Conf. on Intelligent Robots and Systems}, pp. 1170--1176, 2009.

\bibitem{Filgueira.2017}
A.~Filgueira, H.~Gonz{\'a}lez-Jorge, S.~Lag{\"u}ela, L.~D{\'i}az-Vilari{\~n}o,
  and P.~Arias, ``{Quantifying the influence of rain in LiDAR performance},''
  \emph{{Measurement}}, vol.~95, pp. 143--148, 2017.

\bibitem{Hasirlioglu.2017}
S.~Hasirlioglu, I.~Doric, A.~Kamann, and A.~Riener, ``{Reproducible Fog
  Simulation for Testing Automotive Surround Sensors},'' in \emph{{2017 IEEE
  85th Vehicular Technology Conference (VTC Spring)}}, 2017, pp. 1--7.

\bibitem{Hasirlioglu.2016}
S.~Hasirlioglu, I.~Doric, C.~Lauerer, and T.~Brandmeier, ``{Modeling and
  simulation of rain for the test of automotive sensor systems},'' in
  \emph{IEEE Intelligent Vehicle Symposium}, 2016, pp. 286--291.

\bibitem{Hasirlioglu.2016b}
S.~Hasirlioglu, A.~Kamann, I.~Doric, and T.~Brandmeier, ``{Test methodology for
  rain influence on automotive surround sensors},'' in \emph{IEEE International
  Conference on Intelligent Transportation Systems}, 2016, pp. 2242--2247.

\bibitem{Ijaz.2012}
M.~Ijaz, Z.~Ghassemlooy, H.~{Le Minh}, S.~Rajbhandari, and J.~Perez,
  ``{Analysis of fog and smoke attenuation in a free space optical
  communication link under controlled laboratory conditions},'' in \emph{{2012
  International Workshop on Optical Wireless Communications (IWOW)}}, 2012, pp.
  1--3.

\bibitem{Rasshofer.2011}
R.~H. Rasshofer, M.~Spies, and H.~Spies, ``{Influences of weather phenomena on
  automotive laser radar systems},'' \emph{{Advances in Radio Science}},
  vol.~9, pp. 49--60, 2011.

\bibitem{Ryde.2009}
J.~Ryde and N.~Hillier, ``{Performance of laser and radar ranging devices in
  adverse environmental conditions},'' \emph{{Journal of Field Robotics}},
  vol.~26, no.~9, pp. 712--727, 2009.

\bibitem{Sallis.2014}
P.~Sallis, C.~Dannheim, C.~Icking, and M.~Maeder, ``{Air Pollution and Fog
  Detection through Vehicular Sensors},'' in \emph{{2014 8th Asia Modelling
  Symposium}}, 2014, pp. 181--186.

\bibitem{Shamsudin.2016}
A.~U. Shamsudin, K.~Ohno, T.~Westfechtel, S.~Takahiro, Y.~Okada, and
  S.~Tadokoro, ``{Fog removal using laser beam penetration, laser intensity,
  and geometrical features for 3D measurements in fog-filled room},''
  \emph{{Advanced Robotics}}, vol.~30, no. 11-12, pp. 729--743, 2016.

\bibitem{Phillips.2017}
T.~G. Phillips, N.~Guenther, and P.~R. McAree, ``{When the dust settles: The
  four behaviors of LIDAR in the presence of fine airborne particulates},''
  \emph{{Journal of field robotics}}, vol.~34, no.~5, pp. 985--1009, 2017.

\bibitem{Bijelic2018Benchmark}
M.~Bijelic, T.~Gruber, and W.~Ritter, ``{A Benchmark for Lidar Sensors in Fog:
  Is Detection Breaking Down?}'' in \emph{IEEE Intelligent Vehicle
  Symposium}.\hskip 1em plus 0.5em minus 0.4em\relax IEEE, 2018, pp. 760--767.

\bibitem{Sandner.2013}
V.~Sandner, ``{Development of a test target for AEB systems},'' in \emph{{23rd
  international technical conference on the enhanced safety of vehicles
  (ESV)}}, 2013.

\bibitem{Colomb.2008}
M.~Colomb, K.~Hirech, P.~Andr{\'e}, J.~J. Boreux, P.~Lacote, and J.~Dufour,
  ``{An innovative artificial fog production device improved in the European
  project ``FOG''},'' \emph{{Atmospheric Research}}, vol.~87, no.~3, pp.
  242--251, 2008.

\bibitem{Wojtanowski.2014}
J.~Wojtanowski, M.~Zygmunt, M.~Kaszczuk, Z.~Mierczyk, and M.~Muzal,
  ``{Comparison of 905 nm and 1550 nm semiconductor laser rangefinders'
  performance deterioration due to adverse environmental conditions},''
  \emph{{Opto-Electronics Review}}, vol.~22, no.~3, pp. 183--190, 2014.

\bibitem{Kutila.2016}
M.~Kutila, P.~Pyykonen, W.~Ritter, O.~Sawade, and B.~Schaufele, ``Automotive
  lidar sensor development scenarios for harsh weather conditions,'' in
  \emph{IEEE International Conference on Intelligent Transportation Systems},
  2016, pp. 265--270.

\bibitem{kutila2018automotive}
M.~Kutila, P.~Pyyk{\"o}nen, H.~Holzh{\"u}ter, M.~Colomb, and P.~Duthon,
  ``Automotive lidar performance verification in fog and rain,'' in \emph{IEEE
  International Conference on Intelligent Transportation Systems}.\hskip 1em
  plus 0.5em minus 0.4em\relax IEEE, 2018, pp. 1695--1701.

\bibitem{Zhu.2015}
J.~Zhu, D.~Dolgov, and D.~Ferguson, ``{Methods and systems for detecting
  weather conditions including fog using vehicle onboard sensors},'' 2015.

\bibitem{Valeo.2018}
\BIBentryALTinterwordspacing
Valeo. (2018) {Valeo Scala{\circledR} : a laser scanner for highly automated
  driving}. [Online]. Available:
  \url{\url{https://www.valeo.com/en/valeo-scala/}}
\BIBentrySTDinterwordspacing

\bibitem{VelodyneLidarInc..2017}
{Velodyne Lidar Inc.}, ``{Velodyne Lidar Puck: Real-Time 3D~Lidar Sensor},''
  2017.

\bibitem{yang.2012}
W.~Yang, K.~Wang, and W.~Zuo, ``{Neighborhood Component Feature Selection for
  High-Dimensional Data},'' \emph{{JCP}}, vol.~7, no.~1, pp. 161--168, 2012.

\bibitem{Gotzig.2014}
H.~Gotzig and G.~Geduld, ``{Automotive LIDAR},'' in \emph{{Handbook of Driver
  Assistance Systems: Basic Information, Components and Systems for Active
  Safety and Comfort}}, H.~Winner, S.~Hakuli, F.~Lotz, and C.~Singer,
  Eds.\hskip 1em plus 0.5em minus 0.4em\relax {Springer International
  Publishing}, 2014, pp. 1--20.

\end{thebibliography}

\end{document}